\documentclass[english]{article}

\usepackage[T1]{fontenc}
\usepackage[utf8]{inputenc}
\usepackage[margin=1in]{geometry}
\usepackage{color}
\usepackage{babel}
\usepackage{url}
\usepackage{algorithm2e}
\usepackage{amsmath}
\usepackage{amsthm}
\usepackage{amsfonts}
\usepackage{graphicx}
\usepackage[round]{natbib}
\usepackage{xcolor}
\usepackage{pgfplotstable}
\usepgfplotslibrary{fillbetween}
\usepackage{algorithm2e}
\usepackage{tikz}
\usepackage{pgfplots}
\usetikzlibrary{external}
\usepackage{booktabs}
\usepackage[colorlinks=true, allcolors=blue]{hyperref}

\renewcommand{\cite}[1]{\citep{#1}}

\definecolor{yellow}{RGB}{246,217,67}
\definecolor{otherred}{RGB}{235,110,79}
\definecolor{green}{RGB}{99,201,74}
\definecolor{blue}{RGB}{63,72,204}

\definecolor{sfl}{RGB}{135,41,150}
\definecolor{dfl}{RGB}{0,104,150}
\definecolor{ro}{RGB}{255,147,88}

\tikzstyle{box}=[align=left, draw=black, fill=white]

\renewcommand{\arraystretch}{1.2}

\newcommand{\fabian}[1]{\textcolor{red}{[Fabian: #1]}}

\title{Query Suggestion for Retrieval-Augmented Generation via Dynamic In-Context Learning}

\author{
  Fabian Spaeh\\
  \texttt{f.spaeh@celonis.com}
  \and
  Tianyi Chen\thanks{Work done
  while at Celonis, Inc.}\\
  \texttt{ctianyi7@gmail.com}
  \and
  Chen-Hao Chiang\\
  \texttt{c.chiang@celonis.com}
  \and
  Bin Shen\\
  \texttt{b.shen@celonis.com}
}

\begin{document}

\maketitle

\begin{abstract}
Retrieval-augmented generation
with tool-calling agents (agentic RAG)
has become increasingly powerful
in understanding, processing, and
responding to user queries. However,
the scope of the grounding
knowledge is limited and asking
questions that exceed this scope
may lead to issues like hallucination.
While guardrail frameworks aim
to block out-of-scope
questions~\cite{awsguardrails},
no research has
investigated the question of
suggesting answerable queries
in order to complete the user
interaction.

In this paper, we initiate
the study of query suggestion for
agentic RAG.
We consider the setting where user
questions are not answerable, and
the suggested queries should
be similar to aid the user
interaction.
Such scenarios are
frequent for tool-calling LLMs
as communicating the restrictions
of the tools or the underlying
datasets to the user is difficult,
and adding query suggestions
enhances the interaction with
the RAG agent.
As opposed to traditional
settings for query recommendations
such as in search engines,
ensuring that the suggested queries
are answerable is a major
challenge due to the RAG's
multi-step workflow that demands a
nuanced understanding of the RAG
as a whole, which the executing
LLM lacks.
As such, we introduce robust dynamic
few-shot learning which retrieves
examples from relevant workflows.
We show that our system can be
self-learned, for instance on prior
user queries, and is therefore easily
applicable in practice.
We evaluate our approach on
three benchmark datasets based on two
unlabeled question datasets
collected from real-world user
queries.
Experiments on real-world datasets confirm that our method produces more relevant and answerable suggestions, outperforming few-shot and retrieval-only baselines, and thus enable safer, more effective user interaction with agentic RAG.
\end{abstract}

\maketitle

\section{Introduction}

Retrieval-augmented generation (RAG)
allows large language models
(LLMs) to interact and manipulate
data from external sources via tool calls~
\cite{lewis20rag}.
RAG circumvents well-known limitations
of LLMs such as the inability to expand
their memory after training,
carrying out exact computations,
or their lack of domain-specific knowledge
~\cite{naveed2023, taenzer2022, brown2020}.
Even though breakthroughs in LLMs
that facilitate RAG
are recent, they are already
deployed and enjoy increasing popularity~
\cite{patil23,zhang24,patil24}.
Many LLMs now natively support external
tool-calling functionality, simplifying
the implementation of RAG agents.
In our work, we consider RAG agents
that go beyond
free-text retrieval and support arbitrary
tool calls, which is why they are also
referred to as ``agentic.''
Such a system makes progress towards an answer
via a workflow that is dynamically decided
by the LLM. Such a workflow
can include multiple steps of tool
calls and reasoning, before a final
response is delivered to the user.
%

Agentic RAG lowers the barrier
of entry for interacting
with the underlying data.
On the flip side, this means that
typical users will be unaware of the
underlying tool calls and data representation.
This introduces challenges in the
RAG-user interaction and
queries can often fail:
For instance,
users may ask for information that
is not available,
e.g., required information is
not present in a database.
Queries may also
fail due to technical reasons, e.g.,
when a user wants to search in a column
that is not indexed.
To alleviate such issues, we
propose query suggestion
for agentic RAG. Specifically, we aim to
suggest queries to the user that
fulfill two goals:
\begin{enumerate}
    \item The suggested query should be semantically
        similar to the user query and carry the
        same intent as the original query.
    \item The suggested query should be answerable by
        the RAG.
\end{enumerate}
To the best of our knowledge, this problem
is novel in the context of agentic RAG.
%
%
The closest to our problem is the work of
\citet{tayal24}, which considers query suggestion
for RAG for free-text retrieval. However, since
this does not involve an arbitrary agentic workflow,
it is similar
to traditional work on query suggestion.
Compared to this, we face additional challenges
in our setup on agentic RAG: Queries can pertain
to values that appear in any source (e.g., database)
and the number of queries is much larger, because the
agentic RAG can execute an arbitrary set of workflows.
%
Understanding whether a question can be
translated into an executable workflow
by the agentic RAG is difficult,
mainly due to the disconnect
between the LLM,
the available tools,
and the underlying data.
A priori, the LLM does not know
the available data and executing a workflow
successfully requires nuanced
understanding of the available tools
which is difficult to deliver via
prompt instructions.
Furthermore, in order to produce
a valid response, RAG agents
may undergo several layers
of self-reflection \cite{renze24}
and confidence estimation \cite{li24}.
As these are executed on higher level,
understanding the outcome of a query
without executing it becomes challenging.

\subsection{Our Contributions and Techniques}

We are the first to study the problem of query
suggestion for agentic RAG and make
the following contributions towards
this problem.
%
\begin{enumerate}
    \item We propose a query suggestion method
        based on few-shot learning.
        To enhance the LLM's understanding
        about query answerability, 
        we propose robust dynamic few-shot
        learning, which is a novel method
        to retrieve relevant examples
        and automatically combat hallucination.
        We facilitate retrieval through
        a templating approach, which can be
        understood as reducing a query to
        its workflow.

    \item We show how training examples can be
        self-learned, which makes
        our approach immediately applicable.
        We use the
        RAG itself to answer and
        label queries, and we thus
        do not rely on large sets of
        labeled data. As such, our
        approach is practically applicable
        and can learn from prior
        user queries in an unsupervised way.

    \item We experimentally evaluate our
        approach on three real-world
        benchmark datasets, where we consistently
        outperform
        static few-shot learning as a baseline
        and a retrieval-only approach.
        This also shows that
        static few-shot learning
        without the dynamic
        retrieval of examples does
        not lead to effective query
        suggestion, implying that a mere
        description of tools does not suffice
        for the complex task of query
        suggestion for agentic RAG.
\end{enumerate}
On a conceptual level, we introduce
a distinction between two kinds
of unanswerable queries:
Queries without a workflow
to answer them, and queries for which the
called tools cannot find
the needed information in the
underlying data.

\subsection{Related Work}

\paragraph{Query Suggestion}

Query suggestion
aims to identify a follow-up query
that the user may ask
\cite{cao08, mei08, baezayates04}.
\citet{tayal24} are the first to
study query suggestion. However, their
work does not apply to agentic workflows,
but is instead tailored to free-text
retrieval. Agentic RAG is more
challenging and requires a more
sophisticated retrieval scheme, as
it is much harder to guarantee
answerability due to the multi-step workflow.
Furthermore, we use a form
of transfer learning which allows
our approach to be entirely
self-learning, while the
method of \citet{tayal24} relies
on hand-crafted training examples.
To the best of our knowledge,
this is the only prior work on query
suggestion for RAG, and it is
not in our tool-calling setting.

Recent work on
traditional query suggestions
for search engines
also suggests the use of LLMs
\cite{mustar22, baek24}.
This line of work is on
general query suggestion, with no
focus on similarity and answerability
in the context of a RAG agent.
Perhaps the closest work is on
task-based search, where the query
suggestion aims to pick up on a user's
intended task and tries to suggest queries
that lead to the completion of the
task~\cite{ding18, garigliotti17}.
This line of work aims to suggest
queries that are related but also
aid to achieve the user's goals.
None of these works consider
query suggestion for RAG.
Query suggestion is also a popular tool
for enhancing search for retrieval systems,
in which context it is referred
to as query expansion.
The idea is that multiple similar queries
can enhance the retrieval if it is based
on nearest neighbor search on embeddings
\citep{ooi15, carpineto12}.
Such systems have been employed for RAG,
but only for free-text retrieval, i.e.,
without tool-calling agents~\citep{ma23, chan24}.

\paragraph{LLM Guardrails}
LLM guardrails
limit the scope of a language model
agent to a dedicated purpose or a
pre-defined interaction path.
In the context of RAG, the agent should only
answer questions that pertain to its
dedicated task and are within
the abilities of its tools.
Guardrails typically intercept
communication with the LLM,
and allow to program constraints
and conversation flows
via an intermediate layer
\cite{rebedea23, guardrailsai}.
\citet{inan23} introduce a dataset
and classifier that detects
different kinds of harmful LLM
inputs or outputs.
For a thorough discussion, we
refer the reader to \cite{dong24}.

\paragraph{Query Disambiguation}
Another related line of research is on
question disambiguation, where the
goal is to respond with questions
that clarify the original
query. The goal
is therefore similar to our work,
as it is triggered
from queries that are unclear and
cannot be answered directly.
Large language models have been
used to generate such disambiguating
questions for information retrieval
\cite{zamani20}, for agents that
answer open domain questions
\cite{kuhn22, aliannejadi20, kim23}.
This line of research does
not extend to agentic RAG.

\paragraph{Organization}
In Section \ref{sec:main}, we define
three answerability categories and
give a high-level overview of our approach.
Section~\ref{sec:dynamic-fsl} details
our dynamic few-shot learning approach
and templating.
In Section~\ref{sec:self-learning}, we show
how to use the RAG agent to automatically
label example queries with their answerability.
We follow up with an
experimental evaluation in
Section~\ref{sec:experiments},
where we also detail the deployment.

\section{Query Suggestion for RAG}
\label{sec:main}

In order to guide the query suggestion,
we need to make the following classification
about the \emph{answerability} of a query.
On a high level, we differentiate between
queries that are not answerable due to a
limitation in the agent configuration or available tools,
or due to limitations in the underlying data.
We use the following three categories
to classify queries.
\begin{enumerate}
    \item \emph{No workflow:} The RAG agent
        cannot answer the question as it
        cannot be transformed into an executable
        workflow given the RAG's
        purpose and tools.
        This can be because the question
        is out of scope, or the RAG agent does
        not have access to the necessary
        tools to compute a solution.
    \item \emph{No knowledge:} The RAG agent
        understands the question, and is
        able to translate the question into
        a series of tool calls to answer it.
        However, there is no data pertaining
        to the specific values in the query.
        This appears whenever a tool connecting to
        external knowledge responds with an
        empty or invalid result, which makes
        the final RAG response not meaningful.
        As opposed to the first category, this 
        stems from limitations in the underlying
        data.
    \item \emph{Answerable:} The RAG agent
        understood the question and delivered
        a meaningful answer.
        In this case, the user is
        satisfied with the answer
        and not interested
        in similar questions but would want
        to end the conversation or move on
        to different topics,
        which are not the focus of this work.
\end{enumerate}

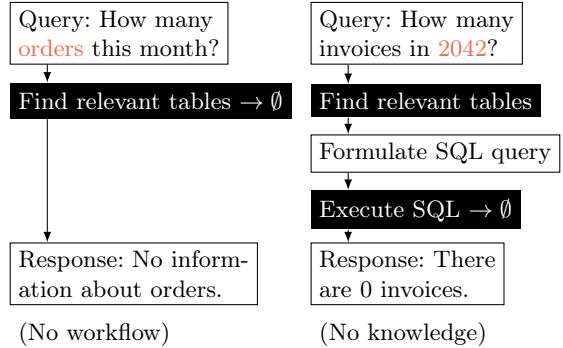
\begin{figure}
    \centering
    \small
    \begin{tikzpicture}
        \node[draw,align=left,anchor=west] (qa) at (0, 0)
            {Query: How many \\ {\color{otherred} orders} this month?};
        \node[fill=black,anchor=west] (sa1) at (0, -0.9)
            {\color{white} Find relevant tables $\to \emptyset$};
        \node[draw,align=left,anchor=west] (ra) at (0, -3.2)
            {Response: No inform-\\ation about orders.};
        \node[align=left,anchor=west] (la) at (0, -4)
            {(No workflow)};

        \draw[-latex] (qa.south -| 0.5, 0) -- (sa1.north -| 0.5, 0);
        \draw[-latex] (sa1.south -| 0.5, 0) -- (ra.north -| 0.5, 0);
        
        \node[draw,align=left,anchor=west] (qb) at (4, 0)
            {Query: How many \\ invoices in {\color{otherred}2042}?};
        \node[fill=black,anchor=west] (sb1) at (4, -0.9)
            {\color{white} Find relevant tables};
        \node[draw,anchor=west] (sb2) at (4, -1.6)
            {Formulate SQL query};
        \node[fill=black,anchor=west] (sb3) at (4, -2.35)
            {\color{white} Execute SQL $\to \emptyset$};
        \node[draw,align=left,anchor=west] (rb) at (4, -3.2)
            {Response: There \\ are 0 invoices.};

        \draw[-latex] (qb.south -| 4.5, 0) -- (sb1.north -| 4.5, 0);
        \draw[-latex] (sb1.south -| 4.5, 0) -- (sb2.north -| 4.5, 0);
        \draw[-latex] (sb2.south -| 4.5, 0) -- (sb3.north -| 4.5, 0);
        \draw[-latex] (sb3.south -| 4.5, 0) -- (rb.north -| 4.5, 0);
        \node[align=left,anchor=west] (la) at (4, -4)
            {(No knowledge)};
    \end{tikzpicture}
    \caption{Two queries that are not answered.
    Black boxes denote a tool execution.
    (Left) the RAG agent does not
    have access to information
    about orders, i.e. it is out of scope and
    there is no workflow that
    will successfully answer the query.
    (Right) The query fails since the final tool
    call returns an empty response due to a typo
    in the year, indicating that there is
    no underlying knowledge about the year 2042.}
    \label{fig:failure}
\end{figure}

We showcase two examples to highlight the
differences between \emph{no workflow} and
\emph{no knowledge} in Figure~\ref{fig:failure}.
%
%
%
Note that this distinction is not
clear-cut as a dynamic workflow can depend
on the results of prior tool-calls.
Furthermore, some queries demand complex
workflows that the LLM may not be
able to deduce, so even though there is
a possible workflow, it cannot be found.
For simplicity,
we will ignore such issues as they are
not in the focus of our work.


\paragraph{Our Approach}
The goal of our work is to suggest similar
queries to user questions which fall into the
first two answerability categories, and for
which the user would want to follow up with a
corrected query.
In our approach, we first focus
on the workflow of a query and
ignore the specific values that
appear in it. If a query is not
answerable because there is no
workflow to execute it, we
then generate similar queries
that can be executed. Here, we
explicitly ignore values in the
queries via templating, which
we explain in the next
section.
The result is a query that
can be answered if we put in
sensible values that avoid
issues with the data.
In a second step, we therefore
impute values into the query
that make it fully answerable,
while trying to stay similar
to the original query.

\begin{figure*}
    \centering
    \footnotesize
    \begin{tikzpicture}
        \node[draw,anchor=west,align=left] (query) at (-0.5, 0)
            {Query: How many orders\\ are due in September?};
        \node[draw,anchor=west] (values) at (5.25, 0.5)
            {Values: timespan = 2024-09-01 to 2024-09-30};
        \node[draw,anchor=west,align=left] (template) at (5.25, -0.5)
            {Template: How many orders\\ are due in [timespan]?};
        \node[draw,anchor=west,align=left] (positive) at (11.75, -0.16)
            {Positive Examples};
        \node[draw,anchor=west,align=left] (negative) at (11.75, -0.83)
            {Negative Examples};
        \node[draw,dotted,rotate=90, minimum width=2.0cm,align=center] (generation) at (15, -0.16)
            {\emph{Generation}\\ Section~\ref{sec:generation}};

        \draw[-latex] (query) --node[above] {Templating} node[below] {Section~\ref{sec:templating}} (4.75, 0) -- (5, 0.5) -- (values.west);
        \draw[-latex] (4.75, 0) -- (5, -0.5) -- (template.west);
        \draw[-latex] (template.east) --node[above] {Few-Shot Retr.} node[below] {Section~\ref{sec:retrieval}} (11.25, -0.5) -- (11.5, 0 |- positive) -- (positive.west);
        \draw[-latex] (11.25, -0.5) -- (11.5, 0 |- negative) -- (negative.west);
        \draw[-latex] (values) -- (values -| generation.north);
        \draw[-latex] (positive) -- (positive -| generation.north);
        \draw[-latex] (negative) -- (negative -| generation.north);
    \end{tikzpicture}

    \vspace{-5pt}
    \caption{Dynamic Few-Shot Learning. Positive examples are labeled queries
    that are answerable; negative examples are queries for which there is
    no workflow.}
    \label{fig:overview}
\end{figure*}
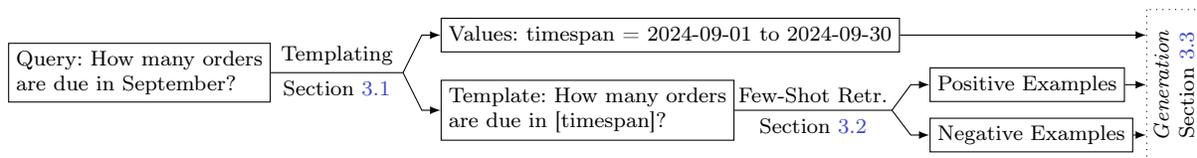

Our basic approach to generate
an answerable query builds on
few-shot learning~\cite{song23}, a
methodology from meta-learning that avoids
retraining the language model by
providing only a few example queries
from the training distribution.
Furthermore, the examples we provide
are not pairs of user queries and
query suggestions. Instead, we transfer-learn
by providing few-shot examples of
answerable and unanswerable queries,
alongside an explanation for the
answerability.
LLMs have shown great ability to
utilize instructions and a few examples
to generalize to unseen
data~\cite{wang22}, and this approach
is also  substantially cheaper than retraining
the model.
For our query suggestion problem, we use
multiple examples that are each annotated
with an explanation, to enhance the
RAGs understanding of answerable queries
and executable workflows.
Specifically, we provide positive and negative
examples. Positive examples are queries that
are classified as answerable, and negative examples
are queries that are not answerable because
they have no workflow.
We ignore queries for which there is
no knowledge since we cannot say with certainty
whether the right combination of values
may make them answerable.
We enhance positive and negative example
queries by providing an explanation why
the RAG agent was able or unable to respond.
This allows the LLM to
understand tools and data better, in order
to generalize better to unseen examples.

Since the LLM is trained on natural
language, it can easily generate similar
queries. However, discerning
possible workflows is difficult and the
answerability of suggested queries
is therefore often low even with
static few-shot
examples, as we show experimentally in
Section~\ref{sec:experiments}.
To overcome this, we want
to provide a much larger set of
positive and negative example queries,
allowing the LLM to infer from
many examples that
capture the nuances of the set of
answerable queries better.
However, the number of examples
we can provide is tightly
limited by the LLM's context size.
We propose a solution that
dynamically retrieves the most
relevant positive and negative example
queries, which we provide as
few-shot examples to the LLM.
Selecting the right
few-shot examples is critical
for in-context learning.
For instance, \citet{agrawal23} 
analyze the performance in translating 
a sentence, while comparing the $n$-gram overlap
of the sentence with the few-shot examples.
Our methodology also avoids
retraining the model,
which is beneficial as retraining
can be expensive and time-consuming.
Furthermore, in
Section~\ref{sec:self-learning},
we show how to use the RAG
to automatically label training examples
and self-learn.

\section{Dynamic Few-Shot Learning}
\label{sec:dynamic-fsl}

For this section, we assume access to
a large set of training examples which
are labeled with their answerability
category and an explanation for why
the category is chosen.
Our goal is to identify
relevant example queries as
few-shot examples and generate
query suggestions based on this.
In the first step, we map a query to a
workflow, which we achieve via a templating
approach. In the second step, we show
how this enables us to retrieve relevant
few-shot examples in a robust manner.
Finally, we show how to use these
few-shot examples to generate
query suggestions.
A complete overview is given in
Figure~\ref{fig:overview}.

\subsection{Question Templating}
\label{sec:templating}

Text embeddings are a standard approach
for semantic retrieval and have shown
success in recommender systems,
providing high recall \cite{abdullahi24}.
Embeddings allow for
fast evaluation of semantic similarity
and nearest-neighbor search,
for instance via the cosine
similarity.\footnote{
The cosine similarity of two embedding
vectors $e, e' \in \mathbb R^n$ is equal
to their normalized inner product
$\langle e, e' \rangle / (\| e \|_2 \cdot \|e' \|_2)$.
For this paper, we will assume
that embedding vectors are already normalized,
i.e. $\| e \|_2 = \|e' \|_2 = 1$.
}
Generating queries that have an executable
workflow requires relevant few-shot examples. The
relevance is therefore not dependent on the
specific values that appear in a query.
Instead, we want to retrieve queries that
demand similar workflows, some of which
can be executed (positive example) and some of which not
(negative example).
Whether an entity is a value is agent-specific,
since it depends on the available tools and
the underlying data. Hence, retraining an
embedding model for each agent configuration 
is expensive and requires large amounts
of training examples.
We therefore follow a different approach
that avoids fine-tuning and is able to
robustly map a query to a workflow, by
tasking an LLM to mask all entity values.

Our aim is to replace all entity
values by their entity name. Specifically,
we want to replace all entity values that will
become arguments for a tool call, such that
a template simply corresponds to the
workflow that answers the query.
Here is a simple example of our
templating, in which we introduce a mask
for the timespan:
\begin{multline*}
    \textrm{``How many invoices were processed in September 2021?''}
    \\ \longrightarrow
    \textrm{``How many invoices were processed in [timespan]?''}
\end{multline*}
Note that
the RAG workflow stays the same for all
timespans.
As templates do not pertain to specific
values, they cannot have data issues and
are therefore either answerable
or not answerable.

Templating is an easy task for
LLMs, but we need each tool
to describe its arguments, along with the type
and a few example values. We provide this
and instruct the LLM to replace all entity
values with their entity names. We facilitate this
with an instruction prompt and few-shot
examples. 

\subsection{Dynamic Few-Shot Retrieval}
\label{sec:retrieval}

Our goal is to provide relevant few-shot examples
to the LLM that cover the possible workflows
via template queries. However, even though
the space of template queries is much lower
than that of concrete queries, the number of
potential workflows is still large. This
introduces two issues: We cannot enumerate
all possible template questions, and even if
we had access to a complete list of template
questions, we could not possibly provide all
of them to the LLM as this would quickly
exceed the LLM's context size.
We avoid this problem by
retrieving few-shot examples dynamically,
via an embedding of the template query.
This ensures that few-shot examples are
always relevant.

An immediate problem with the previous
approach is
that (almost) identical queries result
in different answers. One reason for this lies
in faulty outputs as a result of
hallucination in complex multi-step
queries \cite{huang23}.
Such responses, which we refer
to as faulty, may easily poison the
examples. To mitigate the effect
of faulty responses, we make our retrieval
approach robust by approximating a local
majority vote among the retrieved examples.

We detail our approach in
Algorithm~\ref{alg:retrieval}.
On a high level,
we first retrieve
few-shot examples that are
relevant to the user query.
We then cluster the retrieved queries
by their similarity
and outputs one query per cluster.
The answerability
of this query is decided based on the
majority. E.g., if more than half of
the queries in a cluster are considered
not answerable, we use this as a label.
Our concrete implementation of this
strategy is as follows.
We denote
with $\widehat e$ the embedding of
the the templated user query.
We use two
thresholds $\theta_{\mathrm{sim}}$
and $\theta_{\mathrm{div}}$. The former
is used to decide the addition of few-shot
examples when comparing with embedding $\widehat e$, the
latter to decide the removal of few-shot
examples that are too similar to each other.
We keep a counter for each example and iterate
through all examples in arbitrary order. Instead
of directly outputting an example $i$ that is
similar to the user query, we search
other examples $j$ with similarity 
that exceeds
$\langle e_i, e_j \rangle \ge
\theta_{\mathrm{div}}$. If such an example exists,
we instead increase or decrease the counter for $j$
if the answerability of $i$ matches the answerability $j$ or not,
respectively.
The rationale behind this is to ensure that
the confidence on the answerability of each few-shot
example is high, and we use the
described procedure to
approximate a majority vote.

\begin{figure}[ht]
\centering
\begin{minipage}{.7\linewidth}
\SetAlgoLined
\begin{algorithm}[H]
\SetKwProg{Fn}{RetrieveExamples}{$(\widehat e, \theta_{\mathrm{sim}}, \theta_{\mathrm{div}})$}{end}
\Fn{}{
$E \gets \{ (e_i, \mathrm{ans}_i, \mathrm{tmpl}_i) \in D :
    \langle \widehat e, e_i \rangle \ge \theta_{\mathrm{sim}} \}$\\
Let $n = |E|$ \\
$c_1 \gets 1$ \\
$c_2, \dots, c_n \gets 0$\\
\For{$i = 2, \dots, n$}
{
    Let $j^* \gets \arg\max_{1 \le j \le n \textrm{ with } c_j > 0}
        \langle e_i, e_j \rangle$ \\
    \eIf{$\langle e_i, e_{j^*} \rangle \ge \theta_{\mathrm{div}}$}
    {
        \eIf{$\mathrm{ans}_i = \mathrm{ans}_{j^*}$}
        {
            $c_{j^*} \gets c_{j^*} + 1$ \\
        }{
            $c_{j^*} \gets c_{j^*} - 1$ \\
        }
    }{
        $c_i \gets 1$ \\
    }
}
$E^+ \gets \{ \mathrm{tmpl}_i :
    c_i > 0 \textrm{ and }
    \mathrm{ans}_i = \textrm{\emph{answerable}} \}$ \\
$E^- \gets \{ \mathrm{tmpl}_i :
    c_i > 0 \textrm{ and }
    \mathrm{ans}_i = \textrm{\emph{not answerable}} \}$ \\
\Return{$E^+, E^-$}
}

\medskip
\caption{Robust retrieval of dynamic
few-shot examples. Example templates
$\mathrm{tmpl}_i$ are stored along their
normalized embedding vector $e_i$
and the answerability
$\mathrm{ans}_i \in \{ \textrm{\emph{answerable}},
\textrm{\emph{not answerable}} \}$.
We measure all distances
between template embeddings $e_i$ and $e_j$
using the cosine-similarity
$\langle e_i, e_j\rangle$. Additionally,
we limit the positive examples $E^+$
to the $5$ templates which are most
similar to $\widehat e$, and do the
same with the negative examples $E^-$.}
\label{alg:retrieval}
\end{algorithm}
\end{minipage}
\end{figure}

\begin{figure*}
    \centering
    \small
    \begin{tikzpicture}
        \draw[dashed,fill=black!5] (-0.2, 0.2) rectangle (3.89,-2.8);

        \node[fill=white,draw,align=left,anchor=north west] (qa) at (0, 0)
            {Query: How many \\ {\color{otherred} orders} this month?};
        \node[fill=black,anchor=north west] (sa1) at (0, -1.1)
            {\color{white} Find relevant tables $\to \emptyset$};
        \node[fill=white,draw,align=left,anchor=north west] (ra) at (0, -1.8)
            {Response: No inform-\\ation about orders.};

        \draw[-latex] (qa.south -| 0.5, 0) -- (sa1.north -| 0.5, 0);
        \draw[-latex] (sa1.south -| 0.5, 0) -- (ra.north -| 0.5, 0);

        \node[draw=black,anchor=north west] (template) at (5.5, 0) {Template};
        \node[draw=black,anchor=north west] (embedding) at (5.5, -0.8) {Template Embedding};
        \node[draw=black,anchor=north west,align=left] (eval) at (5.5, -1.8)
            {\emph{Answerability:} No workflow \\
            \emph{Explanation:} No table relating \\ to user query};
        \node[draw,rotate=90, minimum width=2.0cm,align=center] (db) at (11, -1.3)
            {Similarity Database};
        \node[draw,dotted,align=center,anchor=north west] (fsl) at (db.east -| 12.5, 0)
            {\emph{Few-Shot Learning}\\ Section~\ref{sec:dynamic-fsl}};

        \draw[-latex] (3.3, 0 |- eval) --node[above] {Evaluation} node[below] {Section~\ref{sec:ans-eval}} (eval);
        \draw[-latex] (qa.east |- template) -- (template);
        \draw[-latex] (template) -- (template |- embedding.north);
        \draw[-latex] (embedding) -- (embedding -| db.north);
        \draw (template) -- (template -| 9.5, 0) -- (10, 0 |- embedding);
        \draw (eval.north -| 9.5, 0) -- (10, 0 |- embedding);
        \draw[-latex] (db.south |- fsl) -- (fsl);

        \node [align=left,anchor=south west] at (11.25, 0 |- db.west)
            {Store the query if \\ answerability is \emph{no} \\ \emph{workflow} or \emph{answerable}};
    \end{tikzpicture}
    \caption{Simultaneous learning and query suggestion. The LLM
    evaluates the answerability of the query based on the chain of tool-calls
    and response produced by the RAG agent. The evaluation is stored in the
    similarity database using the embedding vector of the templated query.
    At the same time, we use the template query for dynamic retrieval of
    few-shot examples.}
    \label{fig:self-learn}
\end{figure*}
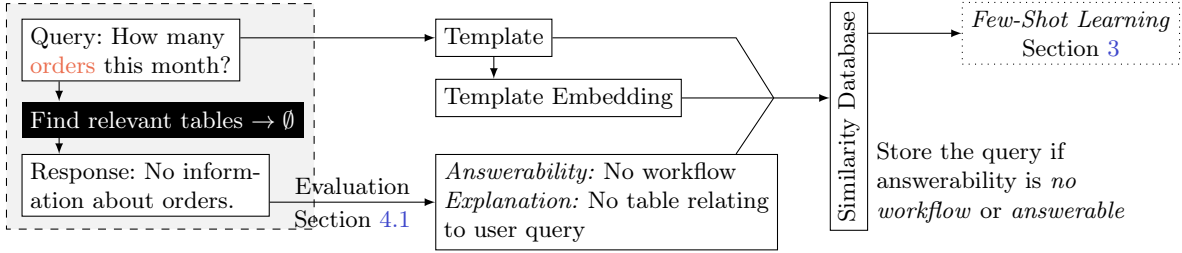

\subsection{Generation}
\label{sec:generation}

\begin{figure}[t]
    \centering
    \small
    \def\arraystretch{1.5}
    \fbox{\begin{tabular}{p{0.9\linewidth}}
        Instructions: You are an agent that suggests queries to
        questions that were not answered. The query ``What is
        the average of days paid late for [timespan]'' was not
        answered. \dots \\
        \midrule
        Positive examples: \\
        -- Template: What are common paths for late payments? \newline
        \phantom{--} Explanation: This was answered because ... \\
        -- Template: How to improve average days paid late during \newline
        \phantom{-- --} [timespan]? \newline
        \phantom{--} Explanation: This was answered because ... \\
        \midrule
        Negative examples: \\
        -- Template: Show average days paid late for [company\! code]. \newline
        \phantom{--} Explanation: Response did not provide specific information. \\
        -- Template: What is the average invoice value in [timespan]? \newline
        \phantom{--} Explanation: The response provided a Python code snippet  \newline
        \phantom{-- --} which does not accurately answer the question. \\
        \midrule
        Task: Generate an answerable template query that is similar
    \end{tabular}}
    \def\arraystretch{1}
    \caption{Prompt for the generation of an answerable
    template query. For brevity, we do not show
    the full prompt and explanations for answerable example queries.
    The explanation for the second negative example gives a strong
    hint that Python code will not be executed, so ad-hoc calculations
    are not allowed for this instance. As a result, we obtain the
    suggestion ``What is the total number of invoices paid late in [timespan]?''
    which can indeed be answered.}
    \label{fig:few-shot}
\end{figure}

Based on the dynamically retrieved
few-shot examples,
we task the LLM with generating
one (or multiple) templates that
serve as answerable templates 
for our query suggestion.
Each template will be converted into
a suggestion, so we obtain
multiple suggested queries.
An example of such a prompt is detailed
in Figure~\ref{fig:few-shot}.

To generate a query suggestion, we
need to replace the masks of
the generated templates with actual
values, making the query fully
answerable without data issues.
For this, we provide
the full history of tool-calls, reasoning
steps, and the final response to the LLM.
We task the LLM to find values for each
mask in this order:
\begin{enumerate}
    \item Consider the values of the original
        query that are used as tool-call arguments.
        If a data issue occurred because of
        a value, the value is ignored.
        Otherwise, use the same value.

    \item Whenever a value is problematic, a tool
        may already provide good alternative values
        in its response. If this is the case,
        use the provided value.

    \item If the value is problematic and the tool
        does not provide alternatives, pick
        example values from the tools.
\end{enumerate}
The steps ensure that the values
picked are as close as possible to the
original query, while keeping the
suggestion answerable.
Note that this requires each tool to provide
good example values from the data it accesses.
Furthermore, programming the tool to suggest
alternative values can improve the suggestion
quality greatly.

\section{Self-Learning}
\label{sec:self-learning}

So far, we assumed the existence
of a large set of labeled queries.
However, to make our dynamic few-shot
approach practically applicable,
we now show that we can use the
RAG agent itself to label queries
with their answerability category
and provide an explanation for the
categorization. As such, our overall
approach requires only a set
of unlabeled queries.
We can learn
directly on user queries, which
additionally avoids a skew between
queries from the training data
and real-world queries.
Figure~\ref{fig:self-learn} shows
our overall self-learning approach
and we show the space of
self-learned examples in
Appendix~\ref{sec:apx}.

There are works that use similar
ideas where the LLM evaluates
its own responses
and this evaluation is used
to inform future decisions.
For instance, \citet{madaan23} suggest to
iteratively improve outputs for a
user question based on LLM feedback.
\citet{shinn23} enable
reinforcement learning through a
reflection module which generates
response feedback via an LLM.
However, combining LLM feedback with our
few-shot retrieval approach
extends the available feedback
beyond the LLMs context size,
which is a challenge for prior work.

\subsection{Response Evaluation}
\label{sec:ans-eval}

It is difficult for an LLM to judge
the answerability of a query up
front as the visible information about
the tools and underlying data is limited.
However, it is possible to determine
the answerability in hindsight after
the RAG agent executed the query.

As apparent from
Figure~\ref{fig:self-learn},
we do this by showing the user query,
the full chain of tool calls and tool
responses, reasoning steps, and the final
RAG response to the LLM.
As such, this is akin to
self-reflection~\cite{rebedea23} but
we do not ask the LLM to judge the
correctness of the answer.
Instead, we task the LLM with
classifying the answerability of
the query.
We facilitate this via a prompt
that contains the instruction
to evaluate the answerability,
along few-shot examples for each
answerability category with
an explanation why the category
applies. We also task the LLM
to output an explanation why
the specific category is chosen.

\section{Experimental Evaluation}
\label{sec:experiments}

We now provide an extensive evaluation
of our approach on three real-world
benchmark datasets
Note that since we are the
first that address the problem of query
suggestion for agentic RAG, there is no
prior work. We thus compare our approach
against  two natural
baselines.
Our results show
that our approach outperforms
both baselines with only
a few thousand unlabeled training examples.
We label the training examples them
with their answerability as described
in Section~\ref{sec:self-learning}.

\subsection{Experimental Setup}

\paragraph{Retrieval-Augmented Generation}
We use a proprietary
query-generating copilot that executes
arbitrary natural language queries
via dynamic workflows.
It operates on top of a knowledge base
akin to a traditional
relational database. This RAG agent is
provided with the database schema,
and queries to retrieve and
display content from the knowledge base.
It offers functionality for
ad-hoc calculations
via a tool that executes
Python code.
The RAG agent
contains a self-reflection module
that evaluates the RAG response
and triggers a re-evaluation
if a mistake in the workflow is detected.
It is built
with GPT-4o~\cite{openai23} in the 2024-08-06
version.
We use three different RAG agents.
Two agents are tasked with
invoice processing,
whereas the first (\textsc{InvoicesNoPython})
does not have access
to the Python tool and the
second (\textsc{InvoicesPython})
does have access
to the Python tool and can therefore
carry out ad-hoc calculations.
The abilities of both agents
therefore differ starkly.
The third agent is tasked with order
management (\textsc{Orders}).

\paragraph{Datasets}
We obtain two datasets from real-world
user logs, recorded over a period of two weeks.
The questions are from related tasks,
but for different RAG agents which
also have different configurations.
This results in $\approx 2000$ questions
for each agent. Because of this, most questions
are related but not unanswerable,
as the RAG agents differ. We
show dataset statistics in
Figure~\ref{fig:dataset-descriptions}.
For our experiments, we report values
from a five-fold cross-validation.

\begin{figure}
    \centering
    \small
    \begin{tabular}{|rllll|}
        \hline
        Name & \# total & no workflow & no knowledge & answerable \\
        \hline
        \textsc{InvoicesNoPython} &
            2029 & 1413 & 225 & 391 \\
        \textsc{InvoicesPython} &
            2029 & 1295 & 246 & 488 \\
        \textsc{Orders} &
            2766 & 384 & 1322 & 1096 \\
        \hline
    \end{tabular}
    \caption{Dataset statistics. We show the number of queries along with their classification, as described
    in Section~\ref{sec:ans-eval}.}
    \label{fig:dataset-descriptions}
\end{figure}

\begin{figure}[ht]
\centering
{\small
\begin{tikzpicture}
  \centering
  \begin{axis}[
        ybar, axis on top,
        title={\textsc{InvoicesNoPython}},
        height=4cm, width=\linewidth,
        bar width=0.35cm,
        ymajorgrids, tick align=inside,
        major grid style={draw=white},
        ymin=0, ymax=100,
        axis x line*=bottom,
        axis y line*=right,
        y axis line style={opacity=0},
        tickwidth=0pt,
        enlarge x limits=0.22,
        legend style={
            at={(0.5,-0.2)},
            anchor=north,
            legend cell align=left,
            legend columns=2,
            /tikz/every even column/.append style={column sep=0.5cm},
            draw=none
        },
        ylabel style={align=center, yshift=-0.7cm},
        ylabel={Percentage (\%) \\ Cosine Similarity ($\cdot 100$)},
        symbolic x coords={
           static few-shot, retrieval-only, (dynamic) few-shot
        },
        xtick=data,
        nodes near coords={
         \pgfmathprintnumber[precision=0]{\pgfplotspointmeta}
        }
    ]
    \addplot [draw=none, fill=otherred!80,
              error bars/.cd, y dir=both, y explicit relative
             ] coordinates {
      (static few-shot, 32.9) +- (0, 0.27)
      (retrieval-only, 34.5) +- (0, 0.20)
      ((dynamic) few-shot, 18.7) +- (0, 0.24)
      };
    \addplot [draw=none,fill=yellow!80,
              error bars/.cd, y dir=both, y explicit relative
             ] coordinates {
      (static few-shot, 12.6) +- (0, 0.5)
      (retrieval-only, 6.9) +- (0, 0.26)
      ((dynamic) few-shot, 17.8) +- (0, 0.22)
      };
    \addplot [draw=none, fill=green!80,
              error bars/.cd, y dir=both, y explicit relative
             ] coordinates {
      (static few-shot, 54.5) +- (0, 0.04)
      (retrieval-only, 58.5) +- (0, 0.09)
      ((dynamic) few-shot, 63.5) +- (0, 0.04)
      };
    \addplot [draw=none, fill=blue!80,
              error bars/.cd, y dir=both, y explicit relative
             ] coordinates {
      (static few-shot, 74) +- (0, 0.08)
      (retrieval-only, 85) +- (0, 0.09)
      ((dynamic) few-shot, 88.5) +- (0, 0.1)
      };

  \end{axis}
\end{tikzpicture}

\medskip

\begin{tikzpicture}
  \centering
  \begin{axis}[
        ybar, axis on top,
        title={\textsc{InvoicesPython}},
        height=4cm, width=\linewidth,
        bar width=0.35cm,
        ymajorgrids, tick align=inside,
        major grid style={draw=white},
        ymin=0, ymax=100,
        axis x line*=bottom,
        axis y line*=right,
        y axis line style={opacity=0},
        tickwidth=0pt,
        enlarge x limits=0.22,
        legend style={
            at={(0.5,-0.2)},
            anchor=north,
            legend cell align=left,
            legend columns=2,
            /tikz/every even column/.append style={column sep=0.5cm},
            draw=none
        },
        ylabel style={align=center, yshift=-0.7cm},
        ylabel={Percentage (\%) \\ Cosine Similarity ($\cdot 100$)},
        symbolic x coords={
           static few-shot, retrieval-only, (dynamic) few-shot
        },
        xtick=data,
        nodes near coords={
         \pgfmathprintnumber[precision=0]{\pgfplotspointmeta}
        }
    ]
    \addplot [draw=none, fill=otherred!80,
              error bars/.cd, y dir=both, y explicit relative
             ] coordinates {
      (static few-shot, 24.9) +- (0, 0.28)
      (retrieval-only, 7.7) +- (0, 0.45)
      ((dynamic) few-shot, 5.9) +- (0, 0.24)
      };
    \addplot [draw=none,fill=yellow!80,
              error bars/.cd, y dir=both, y explicit relative
             ] coordinates {
      (static few-shot, 15.4) +- (0, 0.18)
      (retrieval-only, 13.8) +- (0, 0.41)
      ((dynamic) few-shot, 11.6) +- (0, 0.22)
      };
    \addplot [draw=none, fill=green!80,
              error bars/.cd, y dir=both, y explicit relative
             ] coordinates {
      (static few-shot, 59.7) +- (0, 0.52)
      (retrieval-only, 78.8) +- (0, 0.07)
      ((dynamic) few-shot, 82.5) +- (0, 0.04)
      };
    \addplot [draw=none, fill=blue!80,
              error bars/.cd, y dir=both, y explicit relative
             ] coordinates {
      (static few-shot, 85.4) +- (0, 0.07)
      (retrieval-only, 83.3) +- (0, 0.07)
      ((dynamic) few-shot, 84.7) +- (0, 0.08)
      };

  \end{axis}
\end{tikzpicture}

\medskip

\begin{tikzpicture}
  \centering
  \begin{axis}[
        ybar, axis on top,
        title={\textsc{Orders}},
        height=4cm, width=\linewidth,
        bar width=0.35cm,
        ymajorgrids, tick align=inside,
        major grid style={draw=white},
        ymin=0, ymax=100,
        axis x line*=bottom,
        axis y line*=right,
        y axis line style={opacity=0},
        tickwidth=0pt,
        enlarge x limits=0.22,
        legend style={
            at={(0.5,-0.2)},
            anchor=north,
            legend cell align=left,
            legend columns=2,
            /tikz/every even column/.append style={column sep=0.5cm},
            draw=none
        },
        ylabel style={align=center, yshift=-0.7cm},
        ylabel={Percentage (\%) \\ Cosine Similarity ($\cdot 100$)},
        symbolic x coords={
           static few-shot, retrieval-only, (dynamic) few-shot
        },
        xtick=data,
        nodes near coords={
         \pgfmathprintnumber[precision=0]{\pgfplotspointmeta}
        }
    ]

    \addplot [draw=none, fill=otherred!80,
              error bars/.cd, y dir=both, y explicit relative
             ] coordinates {
      (static few-shot, 92.3) +- (0, 0.07)
      (retrieval-only, 23.2) +- (0, 0.32)
      ((dynamic) few-shot, 24.5) +- (0, 0.45)
      };
    \addplot [draw=none,fill=yellow!80,
              error bars/.cd, y dir=both, y explicit relative
             ] coordinates {
      (static few-shot, 4.6) +- (0, 0.8)
      (retrieval-only, 20.7) +- (0, 0.5)
      ((dynamic) few-shot, 20.3) +- (0, 0.3)
      };
    \addplot [draw=none, fill=green!80,
              error bars/.cd, y dir=both, y explicit relative
             ] coordinates {
      (static few-shot, 3.0) +- (0, 0.9)
      (retrieval-only, 54.8) +- (0, 0.18)
      ((dynamic) few-shot, 55.3) +- (0, 0.27)
      };
    \addplot [draw=none, fill=blue!80,
              error bars/.cd, y dir=both, y explicit relative
             ] coordinates {
      (static few-shot, 85.4) +- (0, 0.05)
      (retrieval-only, 83.7) +- (0, 0.06)
      ((dynamic) few-shot, 92.6) +- (0, 0.04)
      };

    \legend{\ no workflow, \ no knowledge, \ answerable, \ similarity}
  \end{axis}
\end{tikzpicture}
}
\caption{Dynamic few-shot learning produces significantly more answerable and relevant queries across all datasets. We group suggested queries
into three answerability categories using the approach
of Section~\ref{sec:ans-eval} and show the
percentage of queries in each category. Additionally,
we evaluate the cosine similarity between the
user query and the suggestion.
We report mean and standard deviation across
five runs.}
\label{fig:basic-eval}
\end{figure}
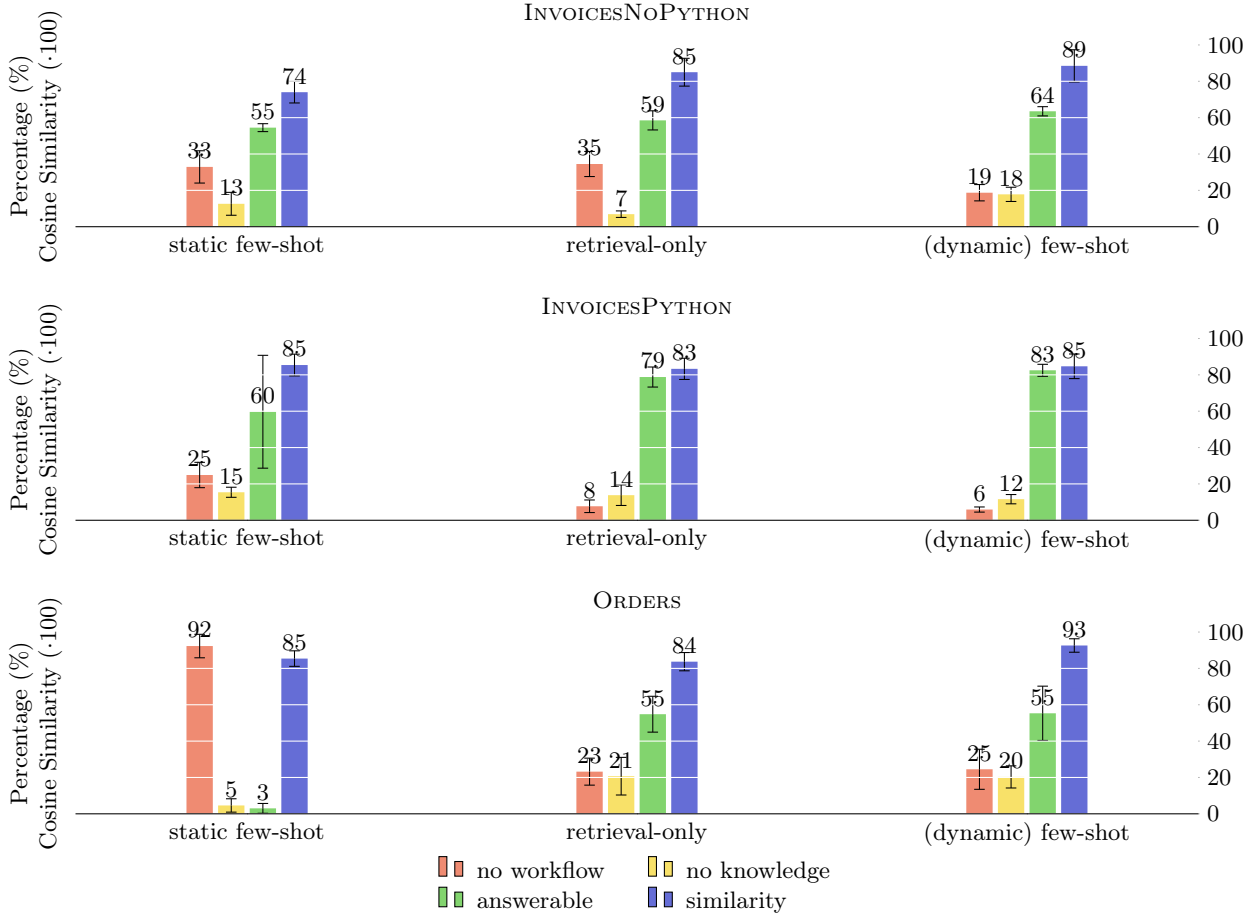

\paragraph{Query Recommendation}
In our evaluation, we use two
natural baselines alongside our approach.
\begin{enumerate}
    \item \emph{Static few-shot}: Query suggestions
        are generated based on instructions and
        a static set of few-shot examples,
        which do not depend on the input query.

    \item \emph{Retrieval-only}: Query suggestions
        are obtained directly from the
        queries retrieved by our robust retrieval approach
        as detailed in Section~\ref{sec:retrieval}
        where only output the most similar positive
        example.
        We merely impute values into
        the retrieved template as in
        Section~\ref{sec:generation}, and task
        the LLM not to change the template.

    \item \emph{Dynamic few-shot}: Our
        method that retrieves examples dynamically
        and uses them as few-shot examples for
        the query suggestion generation,
        as described in Section~\ref{sec:main}.
        We provide at most 5 positive
        and 5 negative examples.
\end{enumerate}

\paragraph{Metrics}
We evaluate both the answerability and the
similarity to the original query. In order
to obtain a score for the answerability
(i.e. not answerable, no knowledge, or
fully answerable), we task the LLM to
evaluate the Copilot's response to the
original query along with all the intermediate
tool calls as described in Section~\ref{sec:ans-eval}.
In order to measure the semantic similarity
between the user query and the suggestion,
we use the cosine-similarity of the embedding
vectors. This is commonly understood as
a good proxy for the semantic similarity 
~\cite{freestone2024, kim2022, mathur2019}.
For simplicity,
we only evaluate answerability and similarity
on the first query suggestion.

\paragraph{Implementation Details}
We use a few practical optimizations to speed
up the query suggestion.
To reduce the number of generated tokens,
when templating a query, we ask the LLM to
provide the list of entity names and
values that appear in a query, instead of asking
to generate the masked sentence. 
We also combine template generation
and value imputation. We ask the LLM to
go through both steps
but not to output the intermediate templates.
For the generation of query suggestions,
we also use GPT-4o~\cite{openai23}. 
For template embeddings, we use
text-embedding-3-small~\cite{embeddings}.

\subsection{Results}

\paragraph{Answerability and Similarity}
We show results for the three benchmark datasets
in Figure~\ref{fig:basic-eval}.
This shows that our dynamic few-shot
method is able to outperform the
two baselines static few-shot learning
and the retrieval-only approach.
As described in the introduction, we
observe that static few-shot learning
has a high trade-off between similarity
and answerability. For instance, on
\textsc{InvoicesPython}, the similarity
matches that of the dynamic few-shot
learning approach, but the answerability
is worse. This is particularly pronounced
on the \textsc{Orders} dataset, where
we can see that the LLM is able to generate
only few answerable queries from
the instructions and few-shot examples
without feedback from the tools.
The retrieval-only approach generally
improves upon the static few-shot
learning. An improvement in answerability
is expected, as suggested queries are
all classified as answerable. However,
on our three benchmark datasets we observe that
the retrieval-only suggestions also have
high similarity to the user queries,
suggesting that the distribution of
user queries is sufficiently narrow.
With dynamic few-shot learning, we
further strictly improve in
terms of similarity
and answerability.
Surprisingly, dynamic few-shot
learning surpasses
the answerability of the
the retrieval-only approach. The latter seems
to be a current barrier due to issues
with hallucination, and many
queries are challenging such that the
RAG agent cannot consistently answer them.
A reason why dynamic few-shot learning
can nonetheless outperform the retrieval-only
approach is that after seeing
relevant positive and negative few-shot
examples, the LLM decides
on more conservative questions which
can be answered with higher certainty.

\paragraph{Answerability-Similarity Tradeoff}
We show the detailed trade-off
between answerability and similarity
for individual suggested queries
in Figure~\ref{fig:sim-ans} in the appendix.
This supports the general observation
from Figure~\ref{fig:basic-eval}
that the answerability and similarity
of our dynamic few-shot learning
approach dominates that
of static few-shot learning and
the retrieval-only approach.
We also see that answerability
of the retrieval-only approach
degrades with increasing similarity.
Dynamic few-shot learning exhibits
little to no tradeoff, suggesting
that it generalizes to
unseen training examples.
The answerability remains
stable for most ranges of similarity,
except for a drop-off for query
suggestions that are very dissimilar. 

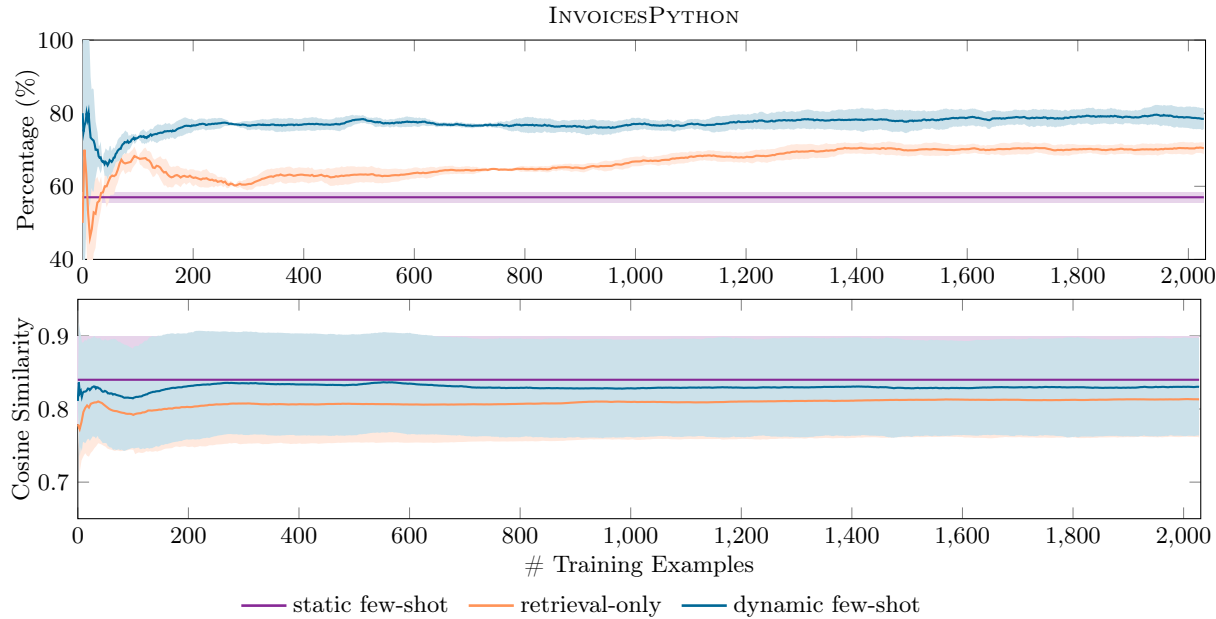
\begin{figure}[ht]
\centering
\small
\begin{tikzpicture}
    \begin{axis}[
    width=\linewidth, 
    height=4.5cm,
    title={\textsc{InvoicesPython}},
    title style={yshift=-3pt},
    xmin=0, xmax=2031,
    ymin=40, ymax=100,
    ylabel style={yshift=-0.5cm},
    xlabel style={yshift=0.15cm},
    xlabel={}, 
    ylabel={Percentage (\%)},
    ]

    \addplot[name path=upper, fill=none, draw=none] table[col sep=comma, x=ix, y expr=100 * (\thisrow{ans2avg} + \thisrow{ans2std})]{data/vary-train-size/zero-shot-eval-python.csv};
    \addplot[name path=lower, fill=none, draw=none] table[col sep=comma, x=ix, y expr=100 * (\thisrow{ans2avg} - \thisrow{ans2std})]{data/vary-train-size/zero-shot-eval-python.csv};
    \addplot[sfl!20] fill between[of=lower and upper];
    \addplot+[color=sfl,mark=.,thick] table[col sep=comma, x=ix, y expr=100 * \thisrow{ans2avg}] {data/vary-train-size/zero-shot-eval-python.csv};

    \addplot[name path=upper, fill=none, draw=none] table[col sep=comma, x=ix, y expr=100 * (\thisrow{ans2avg} + \thisrow{ans2std})]{data/vary-train-size/retrieval-only-eval-python.csv};
    \addplot[name path=lower, fill=none, draw=none] table[col sep=comma, x=ix, y expr=100 * (\thisrow{ans2avg} - \thisrow{ans2std})]{data/vary-train-size/retrieval-only-eval-python.csv};
    \addplot[ro!20] fill between[of=lower and upper];
    \addplot+[solid,color=ro,mark=.,thick] table[col sep=comma, x=ix, y expr=100 * \thisrow{ans2avg}] {data/vary-train-size/retrieval-only-eval-python.csv};

    \addplot[name path=upper, fill=none, draw=none] table[col sep=comma, x=ix, y expr=100 * (\thisrow{ans2avg} + \thisrow{ans2std})]{data/vary-train-size/dynamic-few-shot-eval-python.csv};
    \addplot[name path=lower, fill=none, draw=none] table[col sep=comma, x=ix, y expr=100 * (\thisrow{ans2avg} - \thisrow{ans2std})]{data/vary-train-size/dynamic-few-shot-eval-python.csv};
    \addplot[dfl!20] fill between[of=lower and upper];
    \addplot+[color=dfl,mark=.,thick] table[col sep=comma, x=ix, y expr=100 * \thisrow{ans2avg}] {data/vary-train-size/dynamic-few-shot-eval-python.csv};
    \end{axis}
\end{tikzpicture}

\begin{tikzpicture}
    \begin{axis}[
    width=\linewidth, 
    height=4.5cm,
    xmin=0, xmax=2031,
    ymin=0.65, ymax=0.95,
    ylabel style={yshift=-0.5cm},
    xlabel style={yshift=0.15cm},
    xlabel={\# Training Examples},
    legend style={
        at={(0.45,-0.30)},
        column sep=0pt,
        anchor=north,
        legend cell align=left,
        legend columns=3,
        /tikz/every even column/.append style={column sep=5pt},
        draw=none
    },
    ylabel={Cosine Similarity}]

    \addlegendimage{line legend,sfl,line width=1pt}
    \addlegendentry{Fit}
    \addlegendimage{line legend,ro,line width=1pt}
    \addlegendentry{Fit2}
    \addlegendimage{line legend,dfl,line width=1pt}
    \addlegendentry{Fit3}
    \legend{static few-shot,  retrieval-only, dynamic few-shot}

    \addplot[name path=upper, fill=none, draw=none] table[col sep=comma, x=ix, y expr=\thisrow{simavg} + \thisrow{simstd}]{data/vary-train-size/zero-shot-eval-python.csv};
    \addplot[name path=lower, fill=none, draw=none] table[col sep=comma, x=ix, y expr=\thisrow{simavg} - \thisrow{simstd}]{data/vary-train-size/zero-shot-eval-python.csv};
    \addplot[sfl!20] fill between[of=lower and upper];
    \addplot+[color=sfl,mark=.,thick] table[col sep=comma, x=ix, y=simavg] {data/vary-train-size/zero-shot-eval-python.csv};
    
    \addplot[name path=upper, fill=none, draw=none] table[col sep=comma, x=ix, y expr=\thisrow{simavg} + \thisrow{simstd}]{data/vary-train-size/retrieval-only-eval-python.csv};
    \addplot[name path=lower, fill=none, draw=none] table[col sep=comma, x=ix, y expr=\thisrow{simavg} - \thisrow{simstd}]{data/vary-train-size/retrieval-only-eval-python.csv};
    \addplot[ro!20] fill between[of=lower and upper];
    \addplot+[solid,color=ro,mark=.,thick] table[col sep=comma, x=ix, y=simavg] {data/vary-train-size/retrieval-only-eval-python.csv};

    \addplot[name path=upper, fill=none, draw=none] table[col sep=comma, x=ix, y expr=\thisrow{simavg} + \thisrow{simstd}]{data/vary-train-size/dynamic-few-shot-eval-python.csv};
    \addplot[name path=lower, fill=none, draw=none] table[col sep=comma, x=ix, y expr=\thisrow{simavg} - \thisrow{simstd}]{data/vary-train-size/dynamic-few-shot-eval-python.csv};
    \addplot[dfl!20] fill between[of=lower and upper];
    \addplot+[color=dfl,mark=.,thick] table[col sep=comma, x=ix, y=simavg] {data/vary-train-size/dynamic-few-shot-eval-python.csv};
    \end{axis}
\end{tikzpicture}
\vspace{2pt}
\caption{Percentage of answerable query suggestions
and cosine similarity for a increasing number of labeled
training examples on \textsc{InvoicesPython}.
We use a moving average over the last 50 queries and
report mean and standard deviation.
}
\label{fig:vary-train-size}
\end{figure}

\paragraph{Training Set Size and Self-Learning}
Figure~\ref{fig:vary-train-size} shows
the suggestion performance for a
varying number of training examples
on the \textsc{InvoicesPython} instance.
Here, we choose a random
order of the user queries. We generate
a suggested query for every user query
and evaluate its answerability and
similarity to the original user query.
We also self-learn on all previous user queries.
After an initial period
with high variance in similarity and
answerability,
dynamic few-shot learning quickly
improves in answerability, suggesting
that only 500 user queries are
sufficient to effectively discern the
answerability for suggested queries on
our datasets.
The answerability of the
retrieval-only approach also improves. This shows
that the robust retrieval approach of
Section~\ref{sec:retrieval} can
combat hallucinated responses.
The static few-shot approach offers very
high similarity while suffering
low answerability.

\section{Conclusion}

We introduce and address the novel task of query suggestion for agentic retrieval-augmented generation (RAG), focusing on enhancing user interaction when initial queries are unanswerable. Unlike traditional query suggestion, our setting must reason about multi-step workflows and system capabilities without direct access to the agent’s internal knowledge or tools. To tackle this, we developed a robust dynamic few-shot learning framework that retrieves relevant answerability-aware examples via templating and embedding-based retrieval.

Our approach requires no manual labels, enabling practical deployment through self-learning on past user queries. Empirical results across three real-world RAG agents show that our method significantly outperforms both static few-shot and retrieval-only baselines in terms of both semantic similarity and answerability.

More broadly, our framework suggests a scalable path toward more agent-aware prompting: teaching LLMs to reason about tool affordances and system limitations through example-driven conditioning. Beyond query suggestion, our techniques offer building blocks for adaptive interaction, agent debugging, and workflow inspection in complex agentic systems.

\clearpage
\bibliographystyle{plainnat}
\bibliography{references}

\appendix
\vspace{1cm}

\begin{figure*}[ht!]
    \centering
    \begin{tikzpicture}
        \node (tnse) at (0, 0) {\includegraphics[width=8cm]{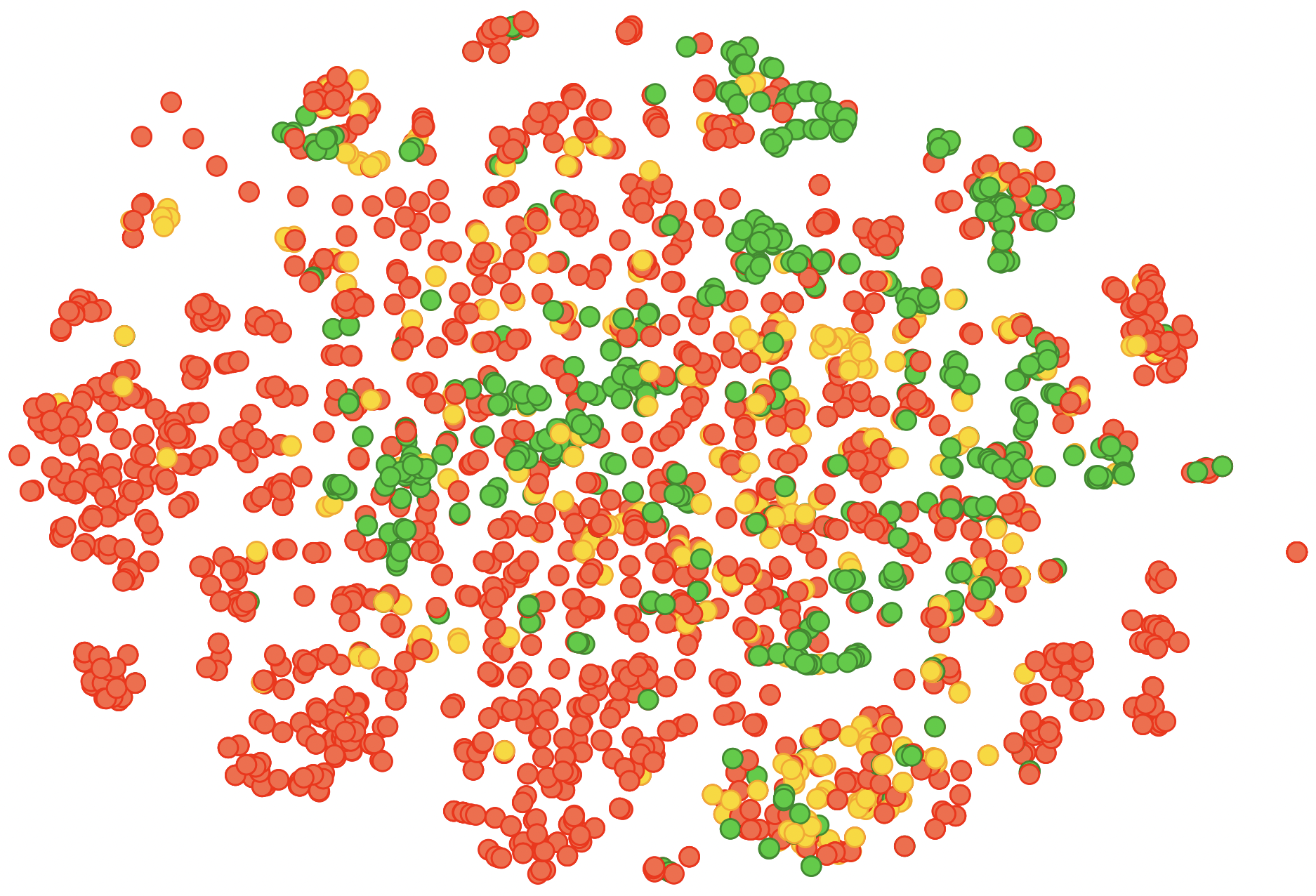}};
        \node[anchor=north,draw=black,inner sep=0.1, fill=white] (tsne-part) at (tnse.north -| 8cm, 0) {\includegraphics[width=10cm]{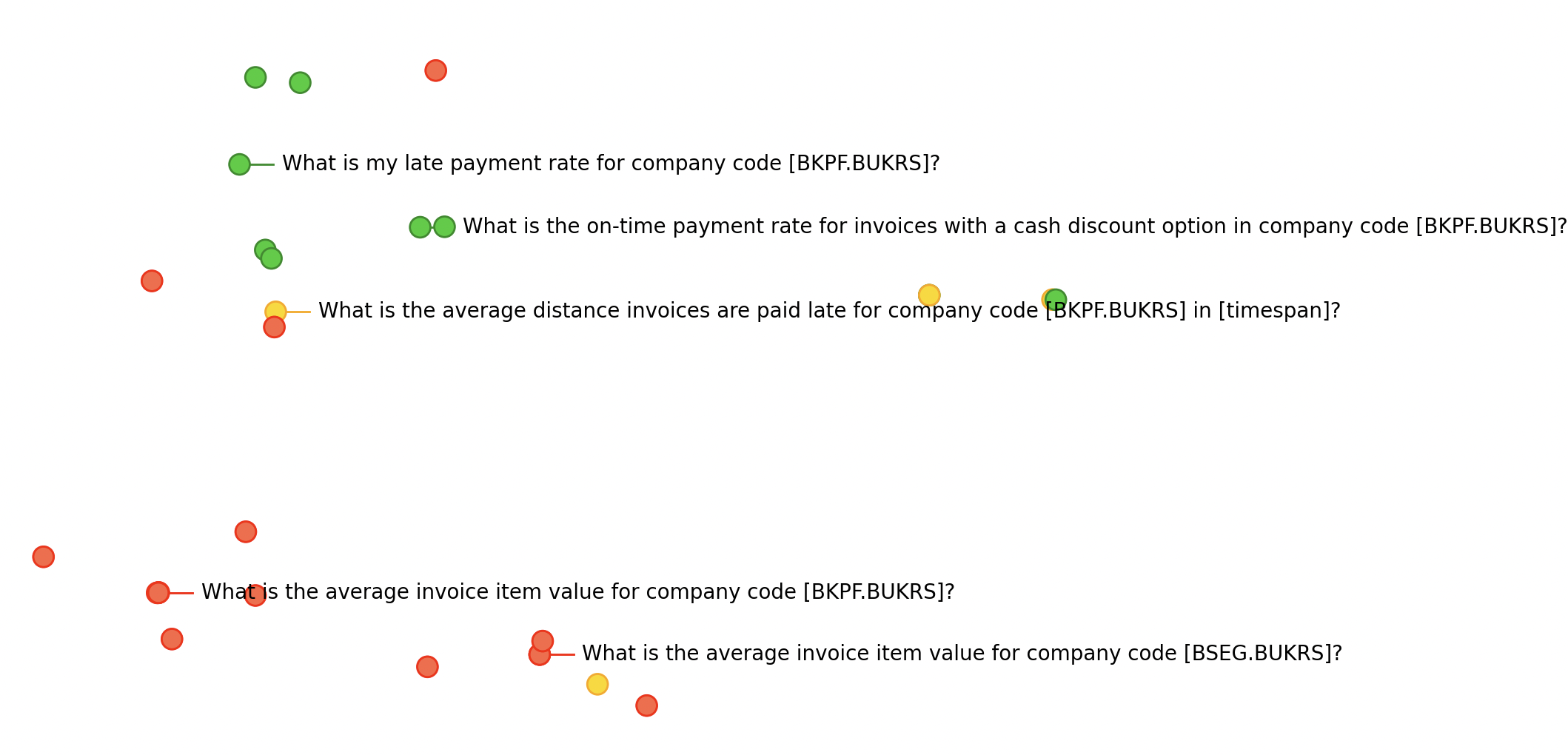}};
        \node[draw=black,inner sep=0.2cm] (zoom) at (2.3cm, -0.25cm) {};
        \draw[] (zoom.north east) -- (tsne-part.north west);
        \draw[] (zoom.south east) -- (tsne-part.south west);
    \end{tikzpicture}
    \caption{Workflow embedding:
    TSNE \cite{maaten14} of template embeddings for
    2029 user questions of the
    \textsc{InvoicesNoPython} dataset.
    Red, yellow, and green dots indicate queries
    that have no workflow, no knowledge, and are
    answerable, respectively.}
    \label{fig:tsne}
\end{figure*}

\section{Workflow Embeddings}
\label{sec:apx}

Figure~\ref{fig:tsne} shows the
``workflow embeddings,''
i.e. the embeddings that we obtained
from templating a set of real-world
user queries as described in
Section~\ref{sec:templating}.
The query answerability is decided by the
LLM itself, which makes up the self-learning
component of our approach.
We can discern groups of
answerable and non-answerable questions.
The zoomed-in region focuses
on user queries about late payments.
Without a Python tool,
ad-hoc calculations such as averages cannot
be evaluated.
This shows that discerning
the answerability of a query
requires a nuanced understanding
of the RAG, but this can be delivered
to the LLM via dynamic
few-shot examples through our
retrieval approach as described
in Section~\ref{sec:retrieval}.

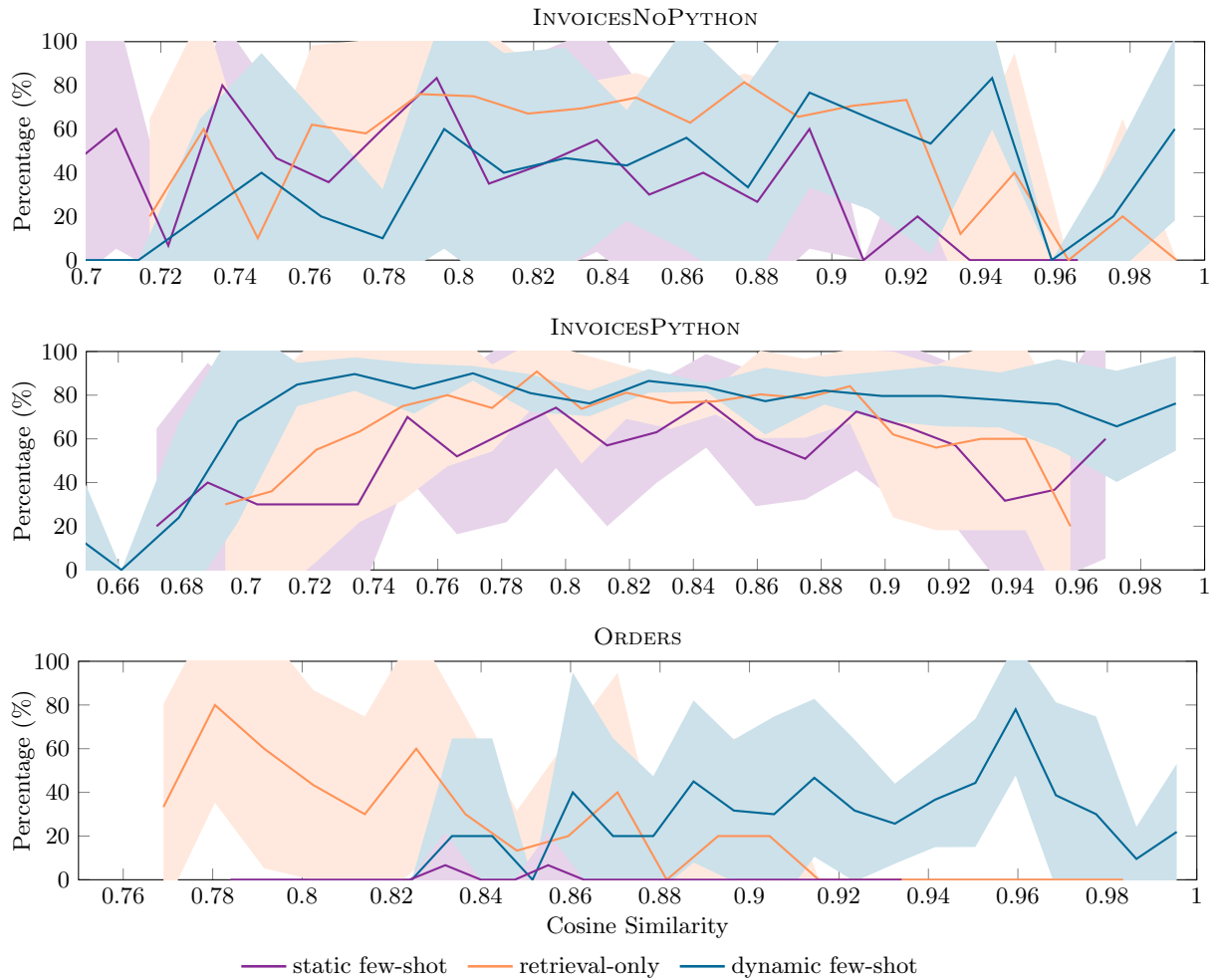
\begin{figure}
\centering
\small
\begin{tikzpicture}
    \begin{axis}[
    title={\textsc{InvoicesNoPython}},
    title style={yshift=-3pt},
    width=\linewidth, 
    height=4.5cm,
    xmin=0.7, xmax=1,
    ymin=0, ymax=100,
    ylabel style={yshift=-0.4cm},
    ylabel={Percentage (\%)},
    legend style={at={(1,0)},anchor=south east,align=left}
    ]

    \addplot[name path=upper, fill=none, draw=none] table[col sep=comma, x=sim, y expr=100 * (\thisrow{ansavg} + \thisrow{ansstd})]{data/ans-vs-sim/static-few-shot-eval-nopython.csv};
    \addplot[name path=lower, fill=none, draw=none] table[col sep=comma, x=sim, y expr=100 * (\thisrow{ansavg} - \thisrow{ansstd})]{data/ans-vs-sim/static-few-shot-eval-nopython.csv};
    \addplot[sfl!20] fill between[of=lower and upper];
    \addplot+[color=sfl,mark=.,thick] table[col sep=comma, x=sim, y expr=100 * \thisrow{ansavg}] {data/ans-vs-sim/static-few-shot-eval-nopython.csv};

    \addplot[name path=upper, fill=none, draw=none] table[col sep=comma, x=sim, y expr=100 * (\thisrow{ansavg} + \thisrow{ansstd})]{data/ans-vs-sim/retrieval-only-eval-nopython.csv};
    \addplot[name path=lower, fill=none, draw=none] table[col sep=comma, x=sim, y expr=100 * (\thisrow{ansavg} - \thisrow{ansstd})]{data/ans-vs-sim/retrieval-only-eval-nopython.csv};
    \addplot[ro!20] fill between[of=lower and upper];
    \addplot+[solid,color=ro,mark=.,thick] table[col sep=comma, x=sim, y expr=100 * \thisrow{ansavg}] {data/ans-vs-sim/retrieval-only-eval-nopython.csv};

    \addplot[name path=upper, fill=none, draw=none] table[col sep=comma, x=sim, y expr=100 * (\thisrow{ansavg} + \thisrow{ansstd})]{data/ans-vs-sim/dynamic-few-shot-eval-nopython.csv};
    \addplot[name path=lower, fill=none, draw=none] table[col sep=comma, x=sim, y expr=100 * (\thisrow{ansavg} - \thisrow{ansstd})]{data/ans-vs-sim/dynamic-few-shot-eval-nopython.csv};
    \addplot[dfl!20] fill between[of=lower and upper];
    \addplot+[color=dfl,mark=.,thick] table[col sep=comma, x=sim, y expr=100 * \thisrow{ansavg}] {data/ans-vs-sim/dynamic-few-shot-eval-nopython.csv};
    \end{axis}
\end{tikzpicture}

\medskip

\begin{tikzpicture}
    \begin{axis}[
    title={\textsc{InvoicesPython}},
    title style={yshift=-3pt},
    width=\linewidth, 
    height=4.5cm,
    xmin=0.65, xmax=1,
    ymin=0, ymax=100,
    ylabel style={yshift=-0.4cm},
    ylabel={Percentage (\%)},
    legend style={at={(1,0)},anchor=south east,align=left}
    ]

    \addplot[name path=upper, fill=none, draw=none] table[col sep=comma, x=sim, y expr=100 * (\thisrow{ansavg} + \thisrow{ansstd})]{data/ans-vs-sim/zero-shot-eval-python.csv};
    \addplot[name path=lower, fill=none, draw=none] table[col sep=comma, x=sim, y expr=100 * (\thisrow{ansavg} - \thisrow{ansstd})]{data/ans-vs-sim/zero-shot-eval-python.csv};
    \addplot[sfl!20] fill between[of=lower and upper];
    \addplot+[color=sfl,mark=.,thick] table[col sep=comma, x=sim, y expr=100 * \thisrow{ansavg}] {data/ans-vs-sim/zero-shot-eval-python.csv};

    \addplot[name path=upper, fill=none, draw=none] table[col sep=comma, x=sim, y expr=100 * (\thisrow{ansavg} + \thisrow{ansstd})]{data/ans-vs-sim/retrieval-only-eval-python.csv};
    \addplot[name path=lower, fill=none, draw=none] table[col sep=comma, x=sim, y expr=100 * (\thisrow{ansavg} - \thisrow{ansstd})]{data/ans-vs-sim/retrieval-only-eval-python.csv};
    \addplot[ro!20] fill between[of=lower and upper];
    \addplot+[solid,color=ro,mark=.,thick] table[col sep=comma, x=sim, y expr=100 * \thisrow{ansavg}] {data/ans-vs-sim/retrieval-only-eval-python.csv};

    \addplot[name path=upper, fill=none, draw=none] table[col sep=comma, x=sim, y expr=100 * (\thisrow{ansavg} + \thisrow{ansstd})]{data/ans-vs-sim/dynamic-few-shot-eval-python.csv};
    \addplot[name path=lower, fill=none, draw=none] table[col sep=comma, x=sim, y expr=100 * (\thisrow{ansavg} - \thisrow{ansstd})]{data/ans-vs-sim/dynamic-few-shot-eval-python.csv};
    \addplot[dfl!20] fill between[of=lower and upper];
    \addplot+[color=dfl,mark=.,thick] table[col sep=comma, x=sim, y expr=100 * \thisrow{ansavg}] {data/ans-vs-sim/dynamic-few-shot-eval-python.csv};
    \end{axis}
\end{tikzpicture}

\medskip

\begin{tikzpicture}
    \begin{axis}[
    title={\textsc{Orders}},
    title style={yshift=-3pt},
    width=\linewidth, 
    height=4.5cm,
    xmin=0.75, xmax=1,
    ymin=0, ymax=100,
    ylabel style={yshift=-0.5cm},
    xlabel style={yshift=0.15cm},
    xlabel={Cosine Similarity},
    ylabel={Percentage (\%)},
    legend style={
        at={(0.45,-0.3)},
        column sep=0pt,
        anchor=north,
        legend cell align=left,
        legend columns=3,
        /tikz/every even column/.append style={column sep=5pt},
        draw=none
    },
    ]

    \addlegendimage{line legend,sfl,line width=1pt}
    \addlegendentry{Fit}
    \addlegendimage{line legend,ro,line width=1pt}
    \addlegendentry{Fit2}
    \addlegendimage{line legend,dfl,line width=1pt}
    \addlegendentry{Fit3}
    \legend{static few-shot,  retrieval-only, dynamic few-shot}

    \addplot[name path=upper, fill=none, draw=none] table[col sep=comma, x=sim, y expr=100 * (\thisrow{ansavg} + \thisrow{ansstd})]{data/ans-vs-sim/retrieval-only-eval-mr.csv};
    \addplot[name path=lower, fill=none, draw=none] table[col sep=comma, x=sim, y expr=100 * (\thisrow{ansavg} - \thisrow{ansstd})]{data/ans-vs-sim/retrieval-only-eval-mr.csv};
    \addplot[ro!20] fill between[of=lower and upper];
    \addplot+[solid,color=ro,mark=.,thick] table[col sep=comma, x=sim, y expr=100 * \thisrow{ansavg}] {data/ans-vs-sim/retrieval-only-eval-mr.csv};

    \addplot[name path=upper, fill=none, draw=none] table[col sep=comma, x=sim, y expr=100 * (\thisrow{ansavg} + \thisrow{ansstd})]{data/ans-vs-sim/dynamic-few-shot-eval-mr.csv};
    \addplot[name path=lower, fill=none, draw=none] table[col sep=comma, x=sim, y expr=100 * (\thisrow{ansavg} - \thisrow{ansstd})]{data/ans-vs-sim/dynamic-few-shot-eval-mr.csv};
    \addplot[dfl!20] fill between[of=lower and upper];
    \addplot+[color=dfl,mark=.,thick,solid] table[col sep=comma, x=sim, y expr=100 * \thisrow{ansavg}] {data/ans-vs-sim/dynamic-few-shot-eval-mr.csv};

    \addplot[name path=upper, fill=none, draw=none] table[col sep=comma, x=sim, y expr=100 * (\thisrow{ansavg} + \thisrow{ansstd})]{data/ans-vs-sim/static-few-shot-eval-mr.csv};
    \addplot[name path=lower, fill=none, draw=none] table[col sep=comma, x=sim, y expr=100 * (\thisrow{ansavg} - \thisrow{ansstd})]{data/ans-vs-sim/static-few-shot-eval-mr.csv};
    \addplot[sfl!20] fill between[of=lower and upper];
    \addplot+[color=sfl,mark=.,thick] table[col sep=comma, x=sim, y expr=100 * \thisrow{ansavg}] {data/ans-vs-sim/static-few-shot-eval-mr.csv};
    \end{axis}
\end{tikzpicture}
\vspace{-3pt}
\caption{Similarity and answerability
of the suggested queries. We show the percentage
of queries that are classified as answerable,
grouped by their cosine similarity to
the original query. We report
mean and standard deviation.}
\label{fig:sim-ans}
\end{figure}

\end{document}